\documentclass[conference]{IEEEtran}
\IEEEoverridecommandlockouts
\usepackage{booktabs}
\usepackage{authblk}
\usepackage{makecell}
\usepackage{cite}
\usepackage{amsmath,amssymb,amsfonts}
\usepackage{algorithmic}
\usepackage{graphicx}
\usepackage{textcomp}
\usepackage{xcolor}
\usepackage{multirow}

\def\BibTeX{{\rm B\kern-.05em{\sc i\kern-.025em b}\kern-.08em
    T\kern-.1667em\lower.7ex\hbox{E}\kern-.125emX}}
\begin{document}

\title{Detecting Neurodegenerative Diseases using Frame-Level Handwriting Embeddings}

\author[ ]{Sarah Laouedj}
\author[ ]{Yuzhe Wang}
\author[ ]{Jes\'us Villalba}
\author[ ]{Thomas Thebaud}
\author[ ]{Laureano Moro-Vel\'azquez}
\author[ ]{Najim Dehak}
\affil[ ]{Center for Language and Speech Processing, The Johns Hopkins University, Baltimore, MD, USA}
\affil[ ]{\textit {\{slaoued1, ywang792, jvillal7, tthebau1,  laureano, ndehak3\}}@jhu.edu}

\maketitle

\begin{abstract}

In this study, we explored the use of spectrograms to represent handwriting signals for assessing neurodegenerative diseases, including 42 healthy controls (CTL), 35 subjects with Parkinson’s Disease (PD), 21 with Alzheimer’s Disease (AD), and 15 with Parkinson’s Disease Mimics (PDM). We applied CNN and CNN-BLSTM models for binary classification using both multi-channel fixed-size and frame-based spectrograms. Our results showed that handwriting tasks and spectrogram channel combinations significantly impacted classification performance. The highest F1-score (89.8\%) was achieved for AD vs. CTL, while PD vs. CTL reached 74.5\%, and PD vs. PDM scored 77.97\%. CNN consistently outperformed CNN-BLSTM. Different sliding window lengths were tested for constructing frame-based spectrograms. A 1-second window worked best for AD, longer windows improved PD classification, and window length had little effect on PD vs. PDM.

\end{abstract}


\section{Introduction}
\label{sec:introduction}

Neurodegenerative Diseases (NDs), including Alzheimer's Disease (AD) and Parkinson's Disease (PD), affect millions globally and remain incurable \cite{de2019handwriting}. These diseases stem from genetic mutations, aging, and environmental factors \cite{lamptey2022review}. PD is characterized by reduced dopamine levels, causing tremors and bradykinesia \cite{bartels2009parkinson}, while AD leads to memory loss and cognitive decline due to abnormal protein aggregates \cite{pickett2019amyloid}. Despite the importance of early diagnosis for effective management, current methods such as neuroimaging and clinical evaluations are costly, invasive, and not universally accessible \cite{sharma2015voice}.

Handwriting analysis presents a promising, non-invasive alternative for the early detection of NDs \cite{castrillon2019characterization, impedovo2018dynamic}. Deterioration in motor function, including tremors and reduced fine motor control, directly impacts handwriting quality \cite{thomas2017handwriting}, while cognitive decline affects the structure and coherence of written content \cite{bortolotto2019handwriting}. These impairments often manifest as subtle distortions, such as smaller letter sizes \cite{drotar2013prediction} and changes in kinematic aspects of movement \cite{6910308}, which may serve as early indicators of NDs \cite{singh2021influence}. Consequently, analyzing handwriting patterns offers a cost-effective and accessible means of detecting early-stage motor and cognitive changes associated with these diseases \cite{de2017brief}.

In recent years, machine learning has been increasingly applied to handwriting data for detecting NDs, particularly PD and AD \cite{tuauctan2021artificial}. 
Taleb et al. used spectrograms generated from $x$, $y$, $z$ coordinates, pressure, altitude, azimuth, and timestamps with a CNN, achieving 83.33\% accuracy on the HandPDMultiMC dataset, which includes handwriting samples from PD subjects and healthy controls (CTL). 
A CNN-BLSTM model with data augmentation improved accuracy to 97.62\% \cite{taleb2023detection}. 
Díaz et al. employed 1D convolutions and Bidirectional GRUs, reaching 94.44\% accuracy using spiral drawings from the NewHandPD dataset, which consists of online handwriting signals from PD and CTL subjects \cite{diaz2021sequence}. 
Gil-Martín et al. processed $x$, $y$, $z$ coordinates, pressure, and grip angle with a Fast Fourier Transform, achieving 96.5\% accuracy and 97.7\% F1-score \cite{gil2019parkinson}. 
Naseer et al. fine-tuned AlexNet on handwriting time series, achieving 98.28\% accuracy \cite{naseer2020retracted}. Kamran et al. used AlexNet, VGGNet, and ResNet, reaching 99.22\% accuracy on combined datasets (HandPD, NewHandPD, and Parkinson's Drawing)  \cite{kamran2021handwriting}. 
Wang et al. introduced a Swin Transformer with CycleGAN augmentation, achieving 88.92\% accuracy on HandPD and 92.68\% on NewHandPD \cite{wang2023coordinate}. 
Dao et al. used a 1D-CNN to detect AD, achieving 87.04\% accuracy with augmentation \cite{dao2022detection}, while Cilia et al. employed ResNet50 for AD detection, reaching 81.03\% accuracy \cite{cilia2021online}.

However, the current literature often focuses on one condition at a time—typically PD or AD. Additionally, most research relies on a narrow set of handwriting tasks (e.g., spirals, cursive writing, and simple wave patterns), which may not fully capture the motor and cognitive complexities of NDs. While some studies transform handwriting time series into spectrograms \cite{taleb2023detection, gil2019parkinson}, they often fail to preserve original signal lengths, losing critical temporal information. None of these studies have yet proposed spectrogram decomposition into smaller frames, a technique that could align more easily with other modalities like speech. 

Our study addresses existing gaps by simultaneously investigating AD, PD, and Parkinson's Disease Mimics (PDM)—a group of NDs that closely resembles PD—thereby broadening the diagnostic scope and improving the ability to differentiate between closely related conditions. Using multi-channel frame-based spectrograms with sliding windows, we capture finer temporal and frequency details, enabling future integration with modalities like speech. We also explore various channel combinations for fixed-size spectrograms to identify the most effective inputs for classification. Additionally, we assess the impact of different handwriting tasks on classification performance, identifying the tasks most influential for each ND.

\section{Materials and Methods}
\label{sec:methods}
In this section, we describe the dataset, tasks, preprocessing techniques, and model architectures used in our study. We also outline the experimental setup employed to evaluate the performance of the models.
\subsection{Dataset}

The NLS dataset was collected at the Johns Hopkins Hospital containing the synchronized multimodal signals of 123 individuals performing 14 tasks involving handwriting, speech, and eye-tracking.
In this article, we use only the handwriting data recorded at 250 $Hz$ using a Wacom One 13 tablet, which captures pen pressure, position, and other dynamics \cite{thebaud2024explainable}.
This study categorizes participants into four groups: AD, PD, PDM, and CTL. 
The AD group includes individuals diagnosed with AD and those with Mild Cognitive impairment (MCI) who tested positive for AD biomarkers through fluids or imaging biomarkers. 
The PDM group serves as a neurodegenerative control group for PD, comprising individuals with NDs closely resembling PD.

All participants were examined and provided handwriting data at Johns Hopkins Hospital, with approval from the Johns Hopkins Institutional Review Board. Informed consent was obtained prior to participation. Demographics of the participants are provided in Table \ref{tab:data}.

\subsection{Tasks}

\paragraph{Point Tasks} 
These tasks assess motor stability by asking participants to hold a pen above a point on the tablet, without touching it, for 10 seconds using each hand separately (Point Right, Point Left). Variations include sustaining the pronunciation of the vowel $/ah/$ for more than 5 seconds while performing the task (Point Sustained).

\paragraph{Spiral Tasks} These tasks assess motor control and detect tremor \cite{pereira2018handwritten} by having participants draw spirals either with their right hand (Spiral Right), left hand (Spiral Left), or dominant hand while rapidly repeating the syllables $pa$, $ta$, and $ka$ (Spiral Pataka).

\paragraph{Writing Tasks} These tasks evaluate cognitive and motor skills \cite{letanneux2014micrographia} by having participants copy text (CopyText), write while reading aloud (CopyReadText), engage in spontaneous writing (FreeWrite), and solve arithmetic problems on the tablet while saying the results out loud (Numbers).

\paragraph{Drawing Tasks} These tasks test spatial reasoning, memory, and fine motor skills \cite{cahn2003discrimination} through tasks including copying an image (CopyImage) and subsequently reproducing it from memory (CopyImageMemory), as well as drawing a clock (DrawClock) and a three-dimensional cube (CopyCube).

This study analyzed handwriting signals solely to facilitate a comprehensive analysis and in-depth understanding of this modality before integrating with speech data.

\begin{table}[htbp]
\vspace{-3mm}
\caption{Demographic and clinical characteristics of participants recorded in the dataset, number of participants in each ND group, number and percentage of females, mean age with standard deviation, total number of tasks recorded, and cognitive evaluations using the Montreal Cognitive Assessment (MoCA) ~\cite{julayanont2017montreal} and Movement Disorder Society-Unified Parkinson's Disease Rating Scale III (MDS-UPDRS III)~\cite{movement2003unified} scores.}
\vspace{-3mm}
\begin{center}
\resizebox{\columnwidth}{!}{
\begin{tabular}{|c|c|c|c|c|c|c|}
\hline
\textbf{Category} & \makecell{\textbf{Number of}\\\textbf{Participants}} & \textbf{Females} & \textbf{Age} & \makecell{\textbf{Files}\\\textbf{recorded}} &  \textbf{MoCA} & \textbf{UPDRS3}\\
\hline
CTL	& 42	& 24 (57.1\%)	& 69y ($\pm 11$)	& 757 	& 25.4 	& 17.0 \\
\hline
PD	& 35	& 14 (40.0\%)	& 67y ($\pm 9$)	& 608 	& 25.9 	& 24.3 \\
\hline
PDM	& 15	& 7 (46.7\%)	& 53y ($\pm 14$)	& 283 	& 25.8 	& 21.5 \\
\hline
AD	& 21	& 4 (19.0\%)	& 70y ($\pm 7$)	& 745 	& 19.7 	&  - \\
\hline
Total	& 113	& 49 (43.4\%)	& 66y ($\pm 12$)	& 1840 	&   -	& -  \\
\hline
\end{tabular}
}
\vspace{-5mm}
\label{tab:data}
\end{center}
\end{table}

\subsection{Preprocessing}

\begin{figure*}[htbp]
\vspace{-5mm}
\centerline{\includegraphics[width=0.9\textwidth]{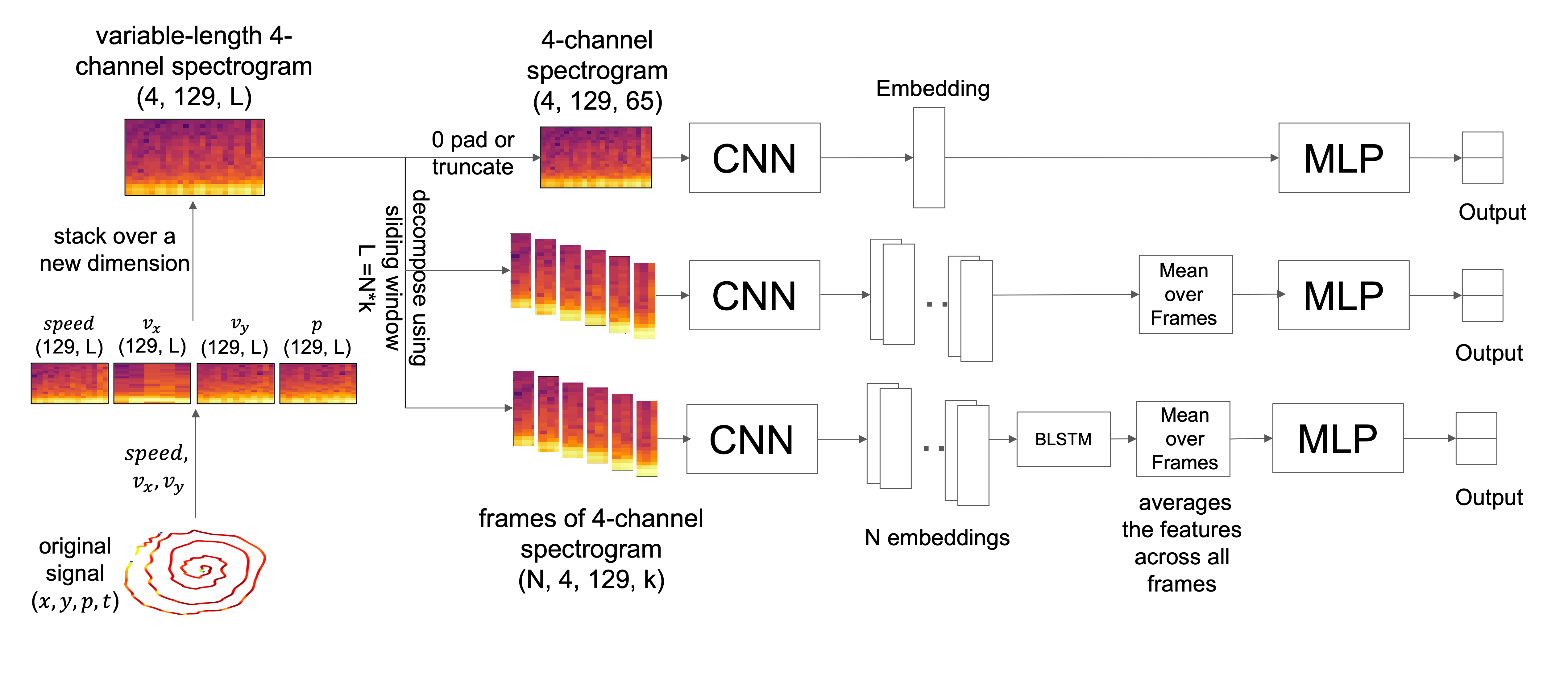}}
\vspace{-5mm}
\caption{Overview of the architectures used in the experiments: The top row represents the CNN architecture with Fixed-Size Spectrograms as input. The middle and bottom rows represent the CNN and CNN-BLSTM architectures using Frame-Based Spectrograms. In the Frame-Based approach, spectrograms were decomposed into smaller frames with a sliding window, and the extracted features were averaged across frames before passing through fully connected layers for classification.}
\label{fig:architectures}
\vspace{-5mm}
\end{figure*}

Figure \ref{fig:architectures} provides a schematic representation of the preprocessing steps. The recorded handwriting signals consist of the pen's $x$ and $y$ coordinates, pressure ($p$), and timestamps ($t$). From these signals, we derived the velocity components (\( v_x = \frac{\Delta x}{\Delta t} \) and \( v_y = \frac{\Delta y}{\Delta t} \)) and calculated $\textit{speed}$ (\(\sqrt{v_x^2 + v_y^2}\)). 

Next, we applied Short-Time Fourier Transform (STFT) with a Blackman window of length 256 and a hop length of 128 to the four signals ($\textit{speed}$, $v_{x}, v_{y}$ and $\textit{pressure}$), generating spectrograms of size (129×$l$) where $l$ is the number of time bins, varying based on the original signal length. We then stacked these spectrograms along a new dimension, creating multi-channel spectrograms of shape (4×129×$l$), where 4  corresponds to the number of input channels (one for each signal).

From there, we explored two preprocessing approaches. The first approach, which we refer to as the \textbf{Fixed-Size Spectrogram Approach}, involved resizing the spectrograms to a fixed size of (4×129×65), achieved by either truncating or zero-padding the time axis for spectrograms longer or shorter than 65-time bins, respectively. This ensured uniform input sizes, with 65-time bins representing the average signal duration of 33 seconds across all data.

In the second approach, which we define as \textbf{Frame-Based Spectrograms Approach}, the original lengths of the spectrograms were preserved, and a sliding window was applied to divide them into smaller frames. The window moved along the time axis with different window sizes ($\textit{25ms, 100ms, 500ms, 1s, 1.5s}$) and strides, extracting non-overlapping frames. If the spectrogram length wasn't divisible by the stride, the final frame was padded to match the window size. This method allowed the model to process the handwriting data in smaller time segments, preserving both temporal and frequency details.

\subsection{Architectures}
Two main architectures were used: a Convolutional Neural Network (CNN) and a CNN-Bidirectional Long Short-Term Memory (CNN-BLSTM) model, both designed to process multi-channel spectrograms derived from handwriting signals.

The CNN architecture consists of a feature extractor and a classifier. The feature extractor has two convolutional layers, each followed by ReLU activation and max-pooling. The first layer uses 32 filters with a 3×3 kernel and a stride of 1, followed by 2×2 max-pooling. The second layer applies 64 filters with the same configuration. The extracted features are flattened and passed to the classifier, which includes two fully connected layers: one with ReLU activation and the final one with softmax for class probabilities. This architecture was tested with both Fixed-Size and Frame-Based Spectrograms. For the latter, the CNN processed each frame separately, and the features were averaged before classification.

The CNN-BLSTM architecture builds on the CNN by adding a BLSTM to capture temporal dependencies. The CNN feature extractor remains the same, but the output is fed into a BLSTM with three layers of 64 hidden units per direction, capturing temporal relationships across frames. The outputs are averaged and passed through a fully connected layer with ReLU activation and a softmax layer for classification. This model was tested only with the Frame-Based Spectrograms.

\subsection{Experiments Setup}
All experiments used 10-fold cross-validation for robust model evaluation. The folds were split by patient to avoid leakage, ensuring each fold included samples from different patients. In each iteration, one fold was used for testing, one for validation, and the rest for training. Models were trained with the Adam optimizer (learning rate 0.001) and cross-entropy loss. Early stopping and a learning rate scheduler (reducing the rate by 0.2 upon validation loss plateau) were applied to prevent overfitting and ensure efficient convergence.

\section{Results and Discussion}
In this section, we report and discuss the results of our experiments for classifying NDs using handwriting spectrograms.

\subsection{Frame-Based Spectrograms with CNN and CNN-BLSTM Models}

To evaluate the effect of varying sliding window lengths on Frame-Based Spectrogram classification across participant groups (AD vs. CTL, PD vs. CTL, and PD vs. PDM), we tested CNN and CNN-BLSTM models using window lengths of 25ms, 100ms, 500ms, 1 second, and 1.5 seconds, as depicted in Figure 2.

For AD vs. CTL, both models performed consistently across window lengths, with the highest F1-score of 81.48\% at the 1-second window. Shorter windows (25ms, 100ms) underperformed due to limited temporal context, while longer windows (1.5 seconds) lowered the performance. 

For PD vs. CTL, the CNN improved with longer windows, peaking at 58.18\% F1-score at 1.5 seconds, likely capturing PD motor patterns. However, CNN-BLSTM struggled with longer windows, peaking at 49.06\% at 100ms, then dropping to 26.67\% at 1.5 seconds, indicating difficulty managing extended temporal data.

For PD vs. PDM, both models performed consistently across all window lengths, with F1-scores between 80\% and 82\%, showing this task is less sensitive to window size.

\begin{figure}[htbp]
\vspace{-3mm}
\centerline{\includegraphics[width=\columnwidth]{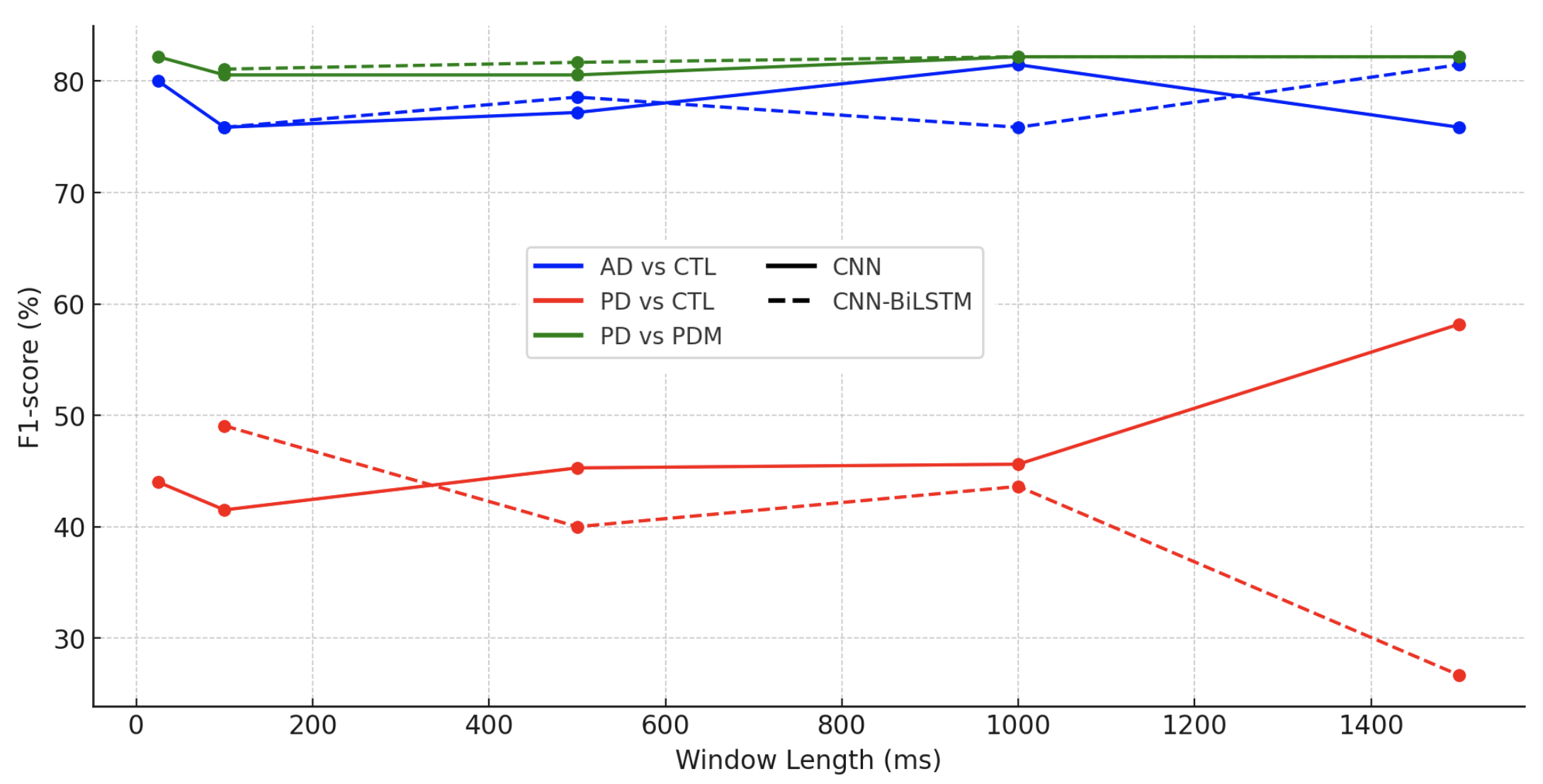}}
\vspace{-3mm}
\caption{F1-score performance of CNN and CNN-BLSTM models using Frame-Based Spectrograms with different window lengths on AD vs CTL, PD vs CTL, and PD vs PDM. The solid lines represent results from the CNN architecture. The dashed lines represent results from the CNN-BLSTM architecture. The window lengths varied from 25ms to 1.5s, with F1-scores plotted for each experiment.}
\vspace{-3mm}
\label{fig: frames_results}
\end{figure}

\subsection{Channels Combination Evaluation with Fixed-Size Spectrograms}
\label{subsection: channels}

In addition to analyzing the effect of window lengths on classification performance, we investigated different combinations of signals derived from handwriting data—specifically $\textit{x}$, $\textit{y}$, pressure($p$), $\textit{trajectory}$ ($\textit{traj}$), $\textit{acceleration}$ ($\textit{acc}$), $\textit{speed}$, $v_{x}$, and $v_{y}$—as input channels for spectrograms to determine the optimal combination for classifying each ND group using Fixed-Size Spectrograms with the CNN model. 

Trajectory was calculated as the cumulative sum of Euclidean distances between consecutive points (\(\sum_{i=1}^{t} \sqrt{(\Delta x_i)^2 + (\Delta y_i)^2}\)), while acceleration was computed as the derivative of velocity (\(\sqrt{{\frac{\Delta v_x}{\Delta t}}^2 + {\frac{\Delta v_y}{\Delta t}}^2}\))

We tested the following signal combinations: \{$x$,$y$,$p$\}, \{\textit{traj},$p$\}, $\{v_{x}, v_{y}, p\}$, $\{\textit{speed}, p\}$, $\{\textit{acc}, p\}$, $\{\textit{traj}, v_{x}, v_{y}, p\}$, $\{\textit{traj}, \textit{acc}, p\}$, $\{\textit{speed}, v_{x}, v_{y}, p\}$, $\{\textit{acc}, v_{x}, v_{y}, p\}$, and $\{\textit{speed}, \textit{acc}, v_{x}, v_{y}, p\}$. Each combination was evaluated to determine the most effective one for classifying each ND group.

For AD vs. CTL, the optimal signal combination was $\{v_{x}, v_{y}, p\}$, achieving an F1-score of 84.62\%, suggesting that Velocity components, capturing directional movement, and pressure provided critical insights into AD-related motor impairments, such as irregular movement patterns and inconsistent grip pressure, distinguishing them from healthy handwriting.

For PD vs. CTL, the best combination was $\{\textit{acc}, v_{x}, v_{y}, p\}$, with an F1-score of 52\%, indicating that acceleration captured tremors and bradykinesia, while velocity and pressure improved the detection of movement dynamics and motor control loss typical in PD.

For PD vs. PDM, $\{\textit{traj}, v_{x}, v_{y}, p\}$ yielded the highest F1-score of 79.41\%, due to trajectory capturing PD-specific tremor patterns, while velocity and pressure highlighting subtle differences in movement speed and control, distinguishing PD from PDM.

\begin{table}[htbp]
\caption{Classification performance metrics for different combinations of signals for Fixed-Size Spectrograms as input to the CNN. The combination $\{v_{x}, v_{y}, p\}$ was used for AD vs CTL, $\{\textit{acc}, v_{x}, v_{y}, p\}$ for PD vs CTL, and $\{\textit{traj}, v_{x}, v_{y}, p\}$ for PD vs PDM. The table reports Accuracy, F1-Score, AUC, Precision, and Recall for each ND pair.}
\begin{center}
\resizebox{\columnwidth}{!}{
\begin{tabular}{|c|c|c|c|c|c|}
\hline
\textbf{Classes} & \textbf{\textit{Accuracy}}& \textbf{\textit{F1-Score}}& \textbf{\textit{AUC}}& \textbf{\textit{Precision}}& \textbf{\textit{Recall}} \\
\hline
AD vs CTL & 84.31 & 84.62 & 86.21 & 73.33 & 100\\
\hline
PD vs CTL & 59.32 & 52 & 59.60 & 65 & 43.33 \\
\hline
PD vs PDM & 68.18 & 79.41 & 55.71 & 71.05 & 90 \\
\hline
\end{tabular}
}
\label{tab: resized spec with CNN}
\vspace{-5mm}
\end{center}
\end{table}

\subsection{Task-Specific Classification Using Optimal Channel Combinations}

We evaluated the CNN model's classification performance on individual handwriting tasks using the optimal signal combinations identified in Subsection \ref{subsection: channels} to generate Fixed-Size multi-channel spectrograms. The results are summarized in Table \ref{tab: tasks}.

For AD vs. CTL, Point Tasks achieved the highest performance, with an F1-score of 89.80\%, particularly in Point Right and Point Left tasks. This highlights the impact of fine motor control impairments in AD, as these tasks require steady hand movements, often compromised in AD patients \cite{kluger1997patterns}.

For PD vs. CTL, performance was moderate, with the best results are for the CopyReadText task, achieving an F1-score of 74.51\%. This reflects the challenge of managing both cognitive and motor functions in PD patients. 

For PD vs. PDM, Point Tasks yielded the best results, with an F1-score of 77.97\%, demonstrating the model's ability to differentiate based on fine motor control. 

\begin{table}[htbp]
\vspace{-3mm}
\caption{Classification results for different handwriting tasks, using CNN with Fixed-Size multi-channel Spectrograms created from different channel combinations. $\{v_{x}, v_{y}, p\}$ for AD vs CTL, $\{\textit{acc}, v_{x}, v_{y}, p\}$ for PD vs CTL, and $\{\textit{traj}, v_{x}, v_{y}, p\}$ for PD vs PDM. The table reports F1-Score and AUC for each task across the three ND pairs.}
\vspace{-3mm}
\vspace{-3mm}
\begin{center}
\resizebox{\columnwidth}{!}{
\begin{tabular}{|c|c|c|c|c|c|c|}
\hline
\multirow{2}{*}{\textbf{Task}} & \multicolumn{2}{c|}{\textbf{AD-CTL}} & \multicolumn{2}{c|}{\textbf{PD-CTL}} & \multicolumn{2}{c|}{\textbf{PD-PDM}} \\
\cline{2-7}
 & \textbf{\textit{F1-Score}} & \textbf{\textit{AUC}} & \textbf{\textit{F1-Score}} & \textbf{\textit{AUC}} & \textbf{\textit{F1-Score}} & \textbf{\textit{AUC}} \\
\hline
Spiral Pataka & 69.09 & 66.26 & 52.46 & 45.88 & 62.96 & 45.74 \\
\hline
Spiral Right & 64.00 & 62.36 & 52.63 & 50.69 & 74.19 & 47.99 \\
\hline
Spiral Left & 44.90 & 43.00 & 62.30 & 56.76 & 72.13 & 49.47 \\
\hline
Spiral Tasks & 69.09 & 66.26 & 55.17 & 52.59 & 74.19 & 51.19 \\
\hline
Point Sustained & 73.47 & 72.91 & 46.81 & 53.03 & 72.73 & 61.41 \\
\hline
Point Right & 84.00 & 84.27 & 52.63 & 51.26 & 67.92 & 57.96 \\
\hline
Point Left & 84.00 & 84.27 & 60.00 & 53.95 & 75.00 & 63.13 \\
\hline 
\textbf{Point Tasks} & \textbf{89.80} & \textbf{90.38} & 55.32 & 61.21 & \textbf{77.97} & \textbf{62.73} \\
\hline
CopyText & 51.16 & 56.48 & 58.62 & 56.48 & 74.58 & 51.79 \\
\hline
\textbf{CopyReadText} & 62.73 & 59.28 & \textbf{74.51} & \textbf{70.54} & 48.65 & 42.26 \\
\hline
FreeWrite & 43.14 & 40.38 & 39.22 & 43.34 & 71.19 & 51.59 \\
\hline
Numbers & 61.90 & 66.98 & 56.67 & 54.31 & 59.65 & 37.50 \\
\hline
Writing Tasks & 59.57 & 62.85 & 32.14 & 35.52 & 63.3 & 42.38\\
\hline
CopyImage & 66.67 & 70.71 & 36.36 & 49.81 & 60.71 & 46.19 \\
\hline
CopyImageMemory & 42.86 & 54.78 & 39.22 & 36.62 & 73.08 & 57.37 \\
\hline
CopyCube & 46.15 & 41.27 & 40.00 & 46.00 & 46.15 & 30.83 \\
\hline
DrawClock & 66.67 & 68.09 & 52.83 & 41.16 & 63.16 & 52.41 \\
\hline
Drawing Tasks & 78.26 & 80.56 & 25.53 & 41.03 & 59.65 & 42.62 \\
\hline
\end{tabular}
}
\label{tab: tasks}
\vspace{-5mm}
\end{center}
\end{table}

\section{Conclusions and Future Work}

In this study, we investigated the use of spectrogram-based embeddings of handwriting signals for the detection of NDs, including AD, PD, and PDM. In summary, the length of the window used to create Frame-Based Spectrograms impacted the results, optimal window length varied by task: 1 second was best for AD vs. CTL using both models, CNN favored longer windows for PD vs. CTL, while CNN-BLSTM preferred shorter windows. Both models handled PD vs. PDM similarly across all window sizes. Additionally, the choice of spectrograms channels significantly affected the performance, velocity, and pressure signals were key in detecting motor impairments, effectively distinguishing AD and PD from control participants and differentiating PD from PDM. Lastly, the results show that different tasks play a key role in distinguishing NDs. Fine motor tasks, like the Point tasks, excelled in differentiating AD from CTL and PD from PDM, while CopyReadText was effective for PD vs. CTL highlighting the value of assessing both cognitive and motor functions. Overall, CNN outperformed CNN-BLSTM, suggesting that the convolutional layers effectively capture the necessary spatial-temporal features, with the added complexity of the BLSTM providing limited additional benefit.

In the future, we plan to increase participant numbers for balanced representation across age, sex, and patient groups. We will explore data augmentation, test different neural network architectures, and consider alternative inputs beyond spectrograms. Additionally, we aim to integrate multimodal data, such as speech and eye tracking, to further enhance classification accuracy across NDs.

\newpage

\bibliographystyle{IEEEtran}
\bibliography{IEEEexample,refs}

\end{document}